\documentclass[11pt,a4paper]{article}
\usepackage[T1]{fontenc}
\usepackage[utf8]{inputenc}
\usepackage{lmodern}
\usepackage[a4paper,margin=2.3cm]{geometry}
\usepackage{graphicx,amsmath,amssymb,booktabs,array,xcolor,makecell}

\usepackage[numbers,sort&compress]{natbib}
\usepackage{authblk}
\usepackage{caption}
\usepackage[hidelinks]{hyperref}
\graphicspath{{figures/}}
\title{BiMa-ECGSR: reconstructing the 12-lead ECG from reduced lead sets with bidirectional state-space models}

\author[1]{Ugo Lomoio}
\author[2]{Pierangelo Veltri\thanks{Corresponding author.}}
\author[1]{Pietro Hiram Guzzi}
\affil[1]{Department of Surgical and Medical Sciences, Magna Graecia University of Catanzaro, Catanzaro, Italy}
\affil[2]{Department of Informatics Modelling and Systems, University of Calabria, Rende, Italy}
\date{}

\begin{document}
\maketitle

\begin{abstract}
Wearables and monitors record few electrocardiogram (ECG) leads; most diagnostic tools assume 12 leads. BiMa-ECGSR reconstructs the 12-lead ECG from reduced lead sets with a bidirectional Mamba state-space encoder--decoder, as regression (BiMa-SR) or diffusion (DiBiMa-SR). Models were trained on PTB-XL for six input sets (I; II; I+II; I+II+V1; I+II+V2; I+II+V1+V4), see only the measured leads at inference, and were tested on the PTB-XL test fold (2,198 records) and two external cohorts (Georgia, 10,275 records; Chapman-Shaoxing/Ningbo, 43,559). From lead II alone, the pooled median Pearson correlation (PCC) over the 11 predicted leads was 0.914 on PTB-XL, against 0.784 for a least-squares linear filter (lead I: 0.906 against 0.789). As limb leads are exactly derivable from I and II, multi-lead inputs were compared on predicted chest leads: 0.939 against 0.849 for I+II and 0.979 against 0.962 for I+II+V1+V4. BiMa-SR exceeded the linear comparator in all 18 configuration--cohort pairs (by 0.012--0.136), a margin that shrank as chest leads were added. From lead II, regression was more accurate than diffusion (DiBiMa-SR: 0.903 one-step, 0.862 with six DDIM steps). A 12-lead classifier trained only on real ECGs reached a superclass macro AUROC of 0.912 on real PTB-XL ECGs, 0.843 on reconstructions from lead II and 0.908 from I+II+V1+V4; externally, reconstructions from three or four leads scored within 0.004 of real ECGs. BiMa-SR (2.68\,M parameters) takes 45.8 ms per 10-s record on a GPU. Reconstruction makes reduced-lead recordings usable by 12-lead tools; it cannot add information absent from the measured leads.
\end{abstract}


\section{Introduction}

The standard 12-lead electrocardiogram (ECG) is the reference recording for most electrocardiographic diagnoses, and its acquisition and interpretation are standardised around the nine-electrode, twelve-lead montage~\cite{kligfield2007recommendations}. Many ECGs, however, are recorded with far fewer leads. Smartwatches record a single lead, usually lead~I, and are used at scale to screen for atrial fibrillation~\cite{perez2019largescale,seshadri2020accuracy}; with repositioning of the watch they can also record other Einthoven and Wilson leads one at a time~\cite{samol2019single,spaccarotella2020multichannel}. Bedside and ambulatory monitors commonly display lead~II for rhythm surveillance, handheld devices provide the six limb leads~\cite{azram2021clinical}, and five-electrode hospital telemetry provides the limb leads plus one precordial lead, typically V1~\cite{drew2004practice,sandau2017update}. Ambulatory Holter and patch monitors record one to three channels over days~\cite{holter1961new,barrett2014comparison,turakhia2013diagnostic,steinberg2017ishne}. The PhysioNet/Computing in Cardiology Challenge 2021 formalised this situation by asking whether reduced lead sets (twelve, six, four, three and two leads) suffice for automated classification~\cite{reyna2021will,reyna2022issues}. Because diagnostic criteria, interpretation software and most deep-learning ECG models~\cite{hannun2019cardiologist,ribeiro2020automatic,attia2019artificial,strodthoff2021deep} assume twelve leads, a reliable reconstruction of the missing leads would let reduced-lead recordings be read by existing 12-lead tools.

Reconstruction of the 12-lead ECG from fewer signals has a long history. The Dower transform derives the 12-lead ECG linearly from the Frank vectorcardiographic leads~\cite{frank1956accurate,dower1980deriving,edenbrandt1988vectorcardiogram}, and the EASI system derives it from four electrodes at non-standard positions~\cite{dower1988deriving,feild2002improved}, with high clinical agreement for rhythm and bundle-branch block~\cite{drew1999accuracy}. Interpolated reduced lead sets record the limb leads and two precordial leads and derive the remaining chest leads by regression~\cite{drew2002comparison,nelwan2004reconstruction,nelwan2004assessment,horacek2008designing}; general (population) coefficients degrade as more precordial leads are removed, whereas patient-specific coefficients need a prior 12-lead calibration recording~\cite{scherer1989synthesis,nelwan2004reconstruction,horacek2008designing}. Reviews summarise these lead systems and synthesis methods~\cite{finlay2007synthesising,tomasic2014electrocardiographic}. Because the limb leads III, aVR, aVL and aVF are exact linear combinations of leads I and II~\cite{einthoven1912different,goldberger1942simple,wilson1934electrocardiograms}, any input set that contains I and II leaves only the precordial leads to be estimated.

Neural networks were first used as committees of multilayer perceptrons on three-lead subsets~\cite{atoui2010novel}; recurrent, convolutional and encoder--decoder networks followed, for chest patches~\cite{sohn2020reconstruction}, ambulatory recorders~\cite{grande2021lead}, single leads~\cite{garg2023single,nakanishi2025deep,samann2026mimonet}, arbitrary lead combinations~\cite{jain2022twelve,kim2025evaluating} and three-lead subsets~\cite{epmoghaddam2025reconstructing,pagotto2025effective,tekin2026lextnet}, as well as additive linear-plus-neural models~\cite{hassannia2025neural}, linear state-space models~\cite{lee2016reconstruction} and studies of electrode selection~\cite{butchy2023importance}. Generative models have also been proposed: generative adversarial networks for single-lead or two-lead input~\cite{seo2022multiple,joo2023twelve,jo2023ecgt2t,zhan2024conditional,presacan2025evaluating}, view synthesis from an inferred cardiac field~\cite{chen2021electrocardio}, signal completion with U-Nets~\cite{lence2024ecgrecover} and denoising diffusion models for missing-lead recovery and lead reconstruction~\cite{cardoso2023bayesian,wei2026condition}, building on diffusion models for ECG generation and time-series imputation~\cite{alcaraz2023diffusion,lai2025diffusets,tashiro2021csdi}. A measurement study of GAN reconstructions from leads I or I+II reported shrunken precordial amplitudes, a regression to the mean~\cite{presacan2025evaluating}.

Four gaps limit how these results can be compared and used. (i) Evaluation designs vary, and it is not always explicit that the model receives only the measured leads at inference and that test cohorts play no part in model selection. (ii) Most studies report waveform agreement on a single dataset; external validation on independently acquired cohorts is rare (cross-dataset results are reported, for example, in~\cite{seo2022multiple}). (iii) Studies usually address one input set, so the benefit of a learned model over linear transforms cannot be followed as leads are added; for input sets containing I and II, aggregate scores over all leads are inflated by limb leads that a linear identity reconstructs exactly. (iv) Few studies ask whether a classifier trained on real 12-lead ECGs can read the reconstructions~\cite{garg2023single,jo2023ecgt2t,wei2026condition}.

Selective state-space models (Mamba~\cite{gu2023mamba}; see also~\cite{gu2022efficiently,dao2024transformers}) process long sequences in linear time, and bidirectional variants scan the sequence in both directions~\cite{zhu2024vision,schiff2024caduceus}; we previously used such a backbone, in regression and diffusion form, for EEG super-resolution~\cite{lomoio2026bidirectional}. Here we introduce \textbf{BiMa-ECGSR}, a bidirectional Mamba encoder--decoder for reduced-lead to 12-lead ECG reconstruction, trained as a regression model (\textbf{BiMa-SR}) and as a diffusion model (\textbf{DiBiMa-SR}). Our contributions are:
\begin{enumerate}
\item One backbone and one training recipe for six clinically motivated input sets: lead I (smartwatch), lead II (monitor), I+II, I+II+V1 (five-electrode telemetry), I+II+V2 and I+II+V1+V4. The models see only the measured leads at inference; the model checkpoint is selected on a fixed validation subset of PTB-XL.
\item An evaluation on the PTB-XL test fold and two external cohorts, Georgia (USA) and Chapman-Shaoxing/Ningbo (China), processed by one identical pipeline and never used for training or model selection, against least-squares linear filters and, for inputs containing I and II, exact Einthoven/Goldberger limb leads; for those inputs the primary metric is restricted to the predicted chest leads.
\item A direct comparison of regression and diffusion from lead II, with one-step and six-step sampling, and a measurement of inference cost on a GPU and a CPU.
\item A diagnostic-readability test in which one classifier trained only on real PTB-XL 12-lead ECGs is applied unchanged to real and reconstructed ECGs of all three cohorts.
\end{enumerate}

\section{Results}

In summary, BiMa-SR reconstructed the predicted chest leads more accurately than the linear comparator in all 18 combinations of input configuration and cohort, with a margin that decreased from 0.136 (lead I, PTB-XL) to 0.016 (I+II+V1+V4, PTB-XL) as measured chest leads were added; from lead II, regression was more accurate than diffusion; and a classifier trained on real ECGs read reconstructions from three or four measured leads with a macro AUROC close to that of real ECGs. The sections below give the evidence in the order of the figures.

\subsection{Data, input configurations and evaluation protocol}

\begin{figure}[p]
\centering
\includegraphics[width=0.82\textwidth]{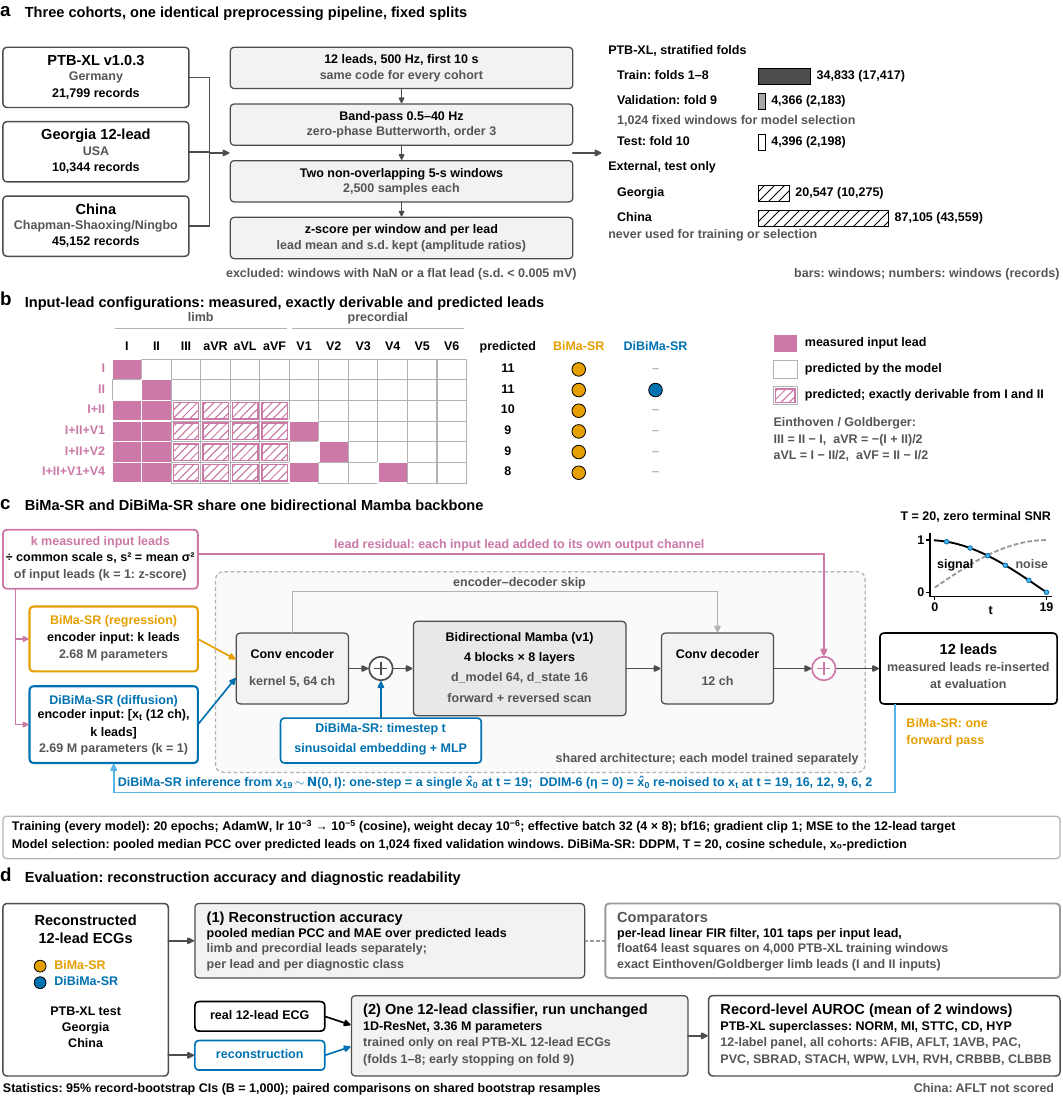}
\caption{\textbf{Architecture and experimental workflow.}
\textbf{a},~Data. Three public 12-lead cohorts (PTB-XL v1.0.3, Germany, 21,799 records; Georgia 12-lead, USA, 10,344 records; Chapman-Shaoxing/Ningbo, China, 45,152 records) pass through one identical pipeline: 12 leads at 500\,Hz, first 10\,s; zero-phase third-order Butterworth band-pass, 0.5--40\,Hz; two non-overlapping 5-s windows (2,500 samples); z-score per window and lead. Windows containing NaN or a flat lead (s.d.\ $<0.005$\,mV) are excluded. PTB-XL folds 1--8: training (34,833 windows, 17,417 records); fold 9: validation (4,366 windows, 2,183 records; a fixed subset of 1,024 windows for model selection); fold 10: test (4,396 windows, 2,198 records). Georgia (20,547 windows, 10,275 records) and China (87,105 windows, 43,559 records) are external test-only cohorts. Bar length $\propto$ number of windows.
\textbf{b},~Input-lead configurations. Pink, measured input leads; white, predicted leads; hatched, predicted leads that are also exactly derivable from I and II (III $=$ II $-$ I, aVR $=-$(I $+$ II)/2, aVL $=$ I $-$ II/2, aVF $=$ II $-$ I/2). Rows: I, II, I+II, I+II+V1 (5-electrode telemetry), I+II+V2, I+II+V1+V4 (11, 11, 10, 9, 9, 8 predicted leads). BiMa-SR is trained for all six configurations; DiBiMa-SR for lead II.
\textbf{c},~Models. BiMa-SR (regression) and DiBiMa-SR (diffusion) share one backbone: a convolutional encoder (kernel 5, 64 channels), a bidirectional Mamba stack (Mamba v1; 4 blocks $\times$ 8 layers; $d_\mathrm{model}$ 64, $d_\mathrm{state}$ 16; forward and reversed scans) and a convolutional decoder (12 output channels) with an encoder--decoder skip. A lead residual adds each input lead to its own output channel; the $k$ inputs share one scale $s$ ($s^2$ = mean input-lead variance). The DiBiMa-SR encoder receives $[x_t$ (12 channels), $k$ input leads$]$ plus a sinusoidal timestep embedding; DiBiMa-SR is a DDPM with $T=20$, a cosine schedule rescaled to zero terminal SNR and $x_0$-prediction, so that sampling starts from pure Gaussian noise. Inset: signal ($\sqrt{\bar\alpha_t}$, solid) and noise ($\sqrt{1-\bar\alpha_t}$, dashed) levels, dots at the DDIM-6 steps. Samplers: one-step (a single $x_0$ prediction at $t=19$) and DDIM-6 ($\eta=0$; $t=19,16,12,9,6,2$). BiMa-SR 2.68\,M, DiBiMa-SR 2.69\,M parameters; training and selection in Methods.
\textbf{d},~Evaluation. (1) Reconstruction accuracy: pooled median PCC and MAE over the predicted leads, for limb and chest leads, per lead and per diagnostic class; comparators: per-lead linear FIR filter (101 taps per input lead, least squares on 4,000 PTB-XL training windows) and, with I and II as inputs, exact Einthoven/Goldberger limb leads. (2) Diagnostic readability: one 12-lead classifier (1D-ResNet, 3.36\,M parameters) trained only on real PTB-XL ECGs (folds 1--8, early stopping on fold 9) is applied unchanged to real and reconstructed ECGs; record-level AUROC on the five PTB-XL superclasses and on a 12-label rhythm/morphology panel shared with Georgia and China (AFLT not scored in China). Statistics: 95\% record-bootstrap CIs ($B=1{,}000$), paired comparisons on shared resamples.}
\label{fig:1}
\end{figure}

Three public 12-lead cohorts were processed by one pipeline (Fig.~\ref{fig:1}a; Methods): PTB-XL (Germany, 21,799 records)~\cite{wagner2020ptbxl}, the Georgia 12-lead database (USA, 10,344 records)~\cite{perezalday2020classification,reyna2021will} and the Chapman-Shaoxing/Ningbo database (China, 45,152 records)~\cite{zheng2020chapman,zheng2020optimal,zheng2022largescale}. Each record contributed two non-overlapping 5-s windows at 500\,Hz after 0.5--40\,Hz band-pass filtering and per-lead z-scoring. The PTB-XL stratified folds 1--8 formed the training set (34,833 windows, 17,417 records), fold 9 the validation set (4,366 windows, 2,183 records; a fixed subset of 1,024 windows was used for model selection) and fold 10 the test set (4,396 windows, 2,198 records). Georgia (20,547 windows, 10,275 records) and China (87,105 windows, 43,559 records) were used only for testing. Quality-control counts are in Supplementary Table~S4.

Six input configurations were studied (Fig.~\ref{fig:1}b): I, II, I+II, I+II+V1, I+II+V2 and I+II+V1+V4, with 11, 11, 10, 9, 9 and 8 predicted leads. BiMa-SR was trained for all six configurations and DiBiMa-SR for lead II. Both share one backbone (Fig.~\ref{fig:1}c): a convolutional encoder, a stack of bidirectional Mamba layers and a convolutional decoder, with a lead residual that adds each input lead to its own output channel; BiMa-SR has 2.68\,M parameters and DiBiMa-SR 2.69\,M. At inference the models receive only the measured leads, and the measured channels of the output are replaced by the true inputs, so that only predicted leads are scored. The comparators were a per-lead linear finite-impulse-response filter (``linear filter'', 101 taps per input lead, least squares on 4,000 PTB-XL training windows) and, for configurations containing I and II, ``Einthoven + linear filter'': exact Einthoven/Goldberger limb leads and the linear filter for the chest leads. Reconstructions were assessed by waveform agreement and by diagnostic readability (Fig.~\ref{fig:1}d).

\subsection{Reconstruction accuracy across input configurations}

\begin{figure}[p]
\centering
\includegraphics[width=\textwidth]{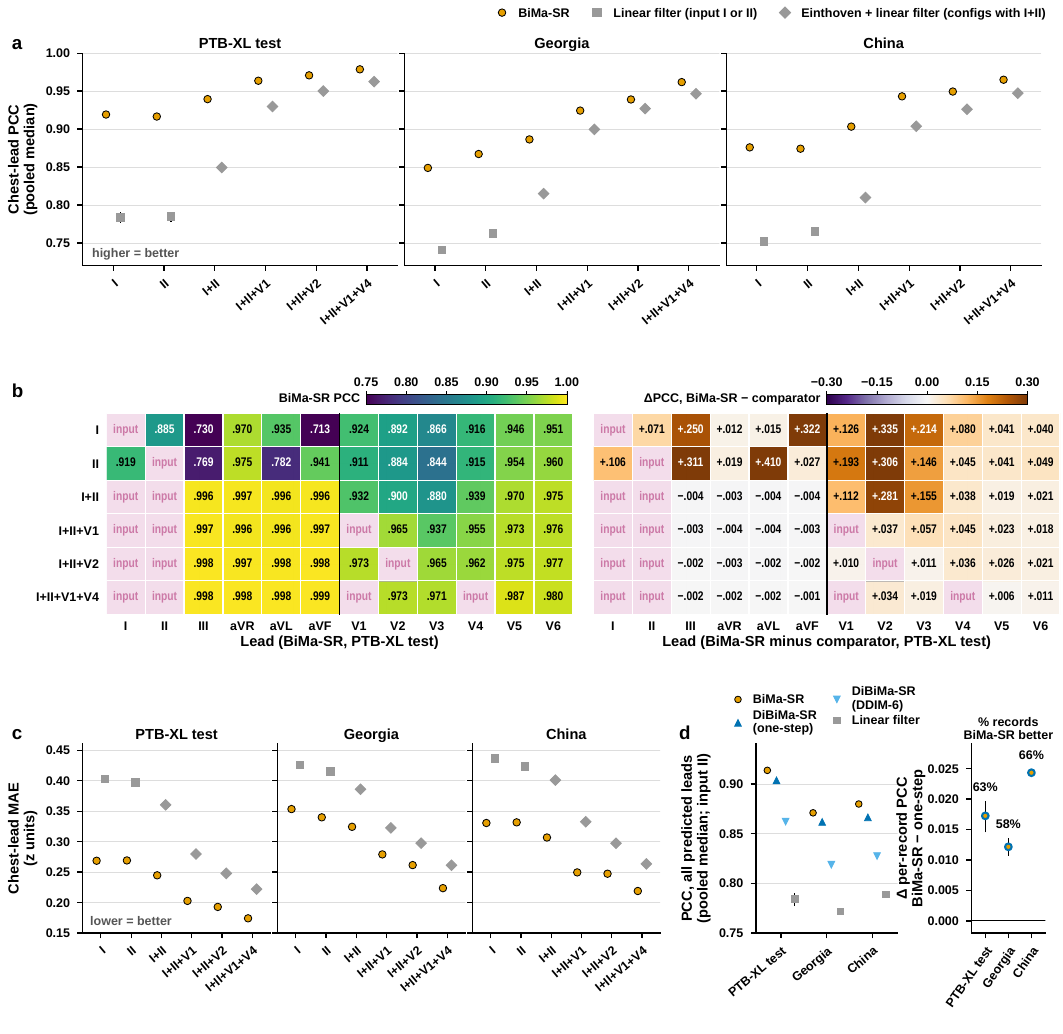}
\caption{\textbf{Reconstruction accuracy across input configurations and cohorts.}
Configurations are ordered by the number of measured leads. Metrics are computed on 5-s windows (500\,Hz, per-lead z-scored) and on predicted leads only. Because III, aVR, aVL and aVF are exactly derivable from I and II, the fair comparison for configurations containing I+II is on the predicted chest leads, the primary metric in \textbf{a} and \textbf{c}. Comparator: linear filter (per-lead FIR, 101 taps per input lead, least squares on 4,000 PTB-XL training windows) for I or II alone; Einthoven + linear filter (exact limb leads, linear filter for chest leads) for configurations containing I and II; on chest leads the two are identical. Cohorts: PTB-XL test 2,198 records / 4,396 windows; Georgia 10,275 / 20,547; China 43,559 / 87,105.
\textbf{a},~Chest-lead pooled median PCC (median over window $\times$ predicted-chest-lead pairs) with 95\% record-bootstrap CIs ($B=1{,}000$; both windows of a record resampled together); most intervals are narrower than the markers. BiMa-SR is higher than the comparator in all 18 configuration--cohort pairs, by 0.012 to 0.136; on PTB-XL test it reaches 0.919 (input I) to 0.979 (I+II+V1+V4), against 0.783 to 0.962.
\textbf{b},~Per-lead pooled median PCC of BiMa-SR on PTB-XL test (left; $n=4{,}396$ windows per cell) and BiMa-SR minus comparator (right; diverging scale centred at 0). Pink, measured input leads (not evaluated). For configurations containing I+II the comparator is exact on limb leads (per-lead PCC $>0.9999$), so the small negative limb differences are expected; for the same reason the all-predicted-lead PCC for I+II+V1+V4 is lower for BiMa-SR (0.995) than for Einthoven + linear filter (0.999), although BiMa-SR is higher on chest leads.
\textbf{c},~Chest-lead MAE (z units; mean over window $\times$ lead pairs) with 95\% record-bootstrap CIs.
\textbf{d},~Lead II input. Left, pooled median PCC over all 11 predicted leads with 95\% CIs for BiMa-SR, DiBiMa-SR (DDPM, $T=20$) sampled in one step or with DDIM-6, and the linear filter. Right, paired per-record PCC difference (per-record PCC $=$ mean over a record's predicted leads and windows), BiMa-SR minus DiBiMa-SR one-step, mean [95\% CI]; label: percentage of records in which BiMa-SR is higher. PTB-XL +0.017 [+0.015, +0.020], 63\%, Wilcoxon $P=3.2\times10^{-48}$; Georgia +0.012 [+0.011, +0.014], 58\%, $P=8.5\times10^{-71}$; China +0.024 [+0.024, +0.025], 66\%, $P<10^{-300}$.}
\label{fig:2}
\end{figure}

From a single limb lead, BiMa-SR reconstructed the 11 missing leads with a pooled median Pearson correlation coefficient (PCC; median over all window $\times$ predicted-lead pairs) of 0.914 [95\% CI 0.911, 0.916] from lead II and 0.906 [0.903, 0.909] from lead I on the PTB-XL test set, against 0.784 and 0.789 for the linear filter (Table~\ref{tab:1}). The mean absolute error (MAE, z units) over the predicted leads was 0.283 against 0.396 from lead II and 0.294 against 0.406 from lead I. The predicted-chest-lead PCC was 0.917 from lead II and 0.919 from lead I, against 0.785 and 0.783 (Fig.~\ref{fig:2}a). The largest per-lead gains were in the leads least correlated with the input (Fig.~\ref{fig:2}b): from lead I, BiMa-SR reached 0.730 in lead III and 0.713 in aVF (linear filter 0.480 and 0.390), and 0.892 and 0.866 in V2 and V3 (linear filter 0.557 and 0.652).

\begin{table}[tbp]
\centering
\caption{\textbf{Reconstruction accuracy per input configuration and cohort.} Pooled median PCC (median over window $\times$ lead pairs) with 95\% record-bootstrap CI ($B=1{,}000$) and mean MAE (z units) of BiMa-SR and of the comparator, over all predicted leads and over the predicted chest leads (measured inputs excluded). Comparator (Comp.): LF, linear filter; E+LF, Einthoven + linear filter (exact limb leads, linear filter for chest leads). For configurations containing I and II the comparator is exact on the limb leads, so the chest-lead columns (bold, BiMa-SR) are the fair comparison. PTB-XL: test fold, 2,198 records; Georgia: 10,275; China: 43,559. Limb-lead values and DiBiMa-SR rows are in Supplementary Table~S1.}
\label{tab:1}
\scriptsize
\setlength{\tabcolsep}{3.2pt}
\begin{tabular}{@{}lll cccc cccc@{}}
\toprule
 & & & \multicolumn{4}{c}{All predicted leads} & \multicolumn{4}{c}{Predicted chest leads} \\
\cmidrule(lr){4-7}\cmidrule(l){8-11}
 & & & \multicolumn{2}{c}{PCC [95\% CI]} & \multicolumn{2}{c}{MAE} & \multicolumn{2}{c}{PCC [95\% CI]} & \multicolumn{2}{c}{MAE} \\
\cmidrule(lr){4-5}\cmidrule(lr){6-7}\cmidrule(lr){8-9}\cmidrule(l){10-11}
Input & Cohort & Comp. & BiMa-SR & Comparator & BiMa & Comp. & BiMa-SR & Comparator & BiMa & Comp. \\
\midrule
I & PTB-XL & LF & \makecell{0.906\\ {[0.903, 0.909]}} & \makecell{0.789\\ {[0.783, 0.795]}} & 0.294 & 0.406 & \makecell{\textbf{0.919}\\ {[0.916, 0.922]}} & \makecell{0.783\\ {[0.776, 0.790]}} & 0.269 & 0.403 \\
 & Georgia & LF & \makecell{0.835\\ {[0.832, 0.837]}} & \makecell{0.754\\ {[0.750, 0.757]}} & 0.371 & 0.426 & \makecell{\textbf{0.849}\\ {[0.846, 0.851]}} & \makecell{0.741\\ {[0.737, 0.745]}} & 0.353 & 0.426 \\
 & China & LF & \makecell{0.857\\ {[0.856, 0.858]}} & \makecell{0.750\\ {[0.748, 0.751]}} & 0.355 & 0.438 & \makecell{\textbf{0.876}\\ {[0.875, 0.877]}} & \makecell{0.752\\ {[0.750, 0.753]}} & 0.331 & 0.436 \\
\midrule
II & PTB-XL & LF & \makecell{0.914\\ {[0.911, 0.916]}} & \makecell{0.784\\ {[0.777, 0.790]}} & 0.283 & 0.396 & \makecell{\textbf{0.917}\\ {[0.914, 0.919]}} & \makecell{0.785\\ {[0.778, 0.791]}} & 0.269 & 0.397 \\
 & Georgia & LF & \makecell{0.871\\ {[0.869, 0.873]}} & \makecell{0.772\\ {[0.768, 0.775]}} & 0.337 & 0.404 & \makecell{\textbf{0.867}\\ {[0.865, 0.869]}} & \makecell{0.763\\ {[0.759, 0.766]}} & 0.340 & 0.415 \\
 & China & LF & \makecell{0.880\\ {[0.879, 0.881]}} & \makecell{0.789\\ {[0.787, 0.790]}} & 0.331 & 0.408 & \makecell{\textbf{0.874}\\ {[0.873, 0.875]}} & \makecell{0.765\\ {[0.764, 0.767]}} & 0.332 & 0.423 \\
\midrule
I+II & PTB-XL & E+LF & \makecell{0.980\\ {[0.980, 0.981]}} & \makecell{0.959\\ {[0.958, 0.961]}} & 0.175 & 0.217 & \makecell{\textbf{0.939}\\ {[0.938, 0.941]}} & \makecell{0.849\\ {[0.845, 0.854]}} & 0.245 & 0.360 \\
 & Georgia & E+LF & \makecell{0.967\\ {[0.967, 0.968]}} & \makecell{0.950\\ {[0.949, 0.951]}} & 0.224 & 0.233 & \makecell{\textbf{0.886}\\ {[0.884, 0.888]}} & \makecell{0.815\\ {[0.812, 0.818]}} & 0.324 & 0.386 \\
 & China & E+LF & \makecell{0.970\\ {[0.969, 0.970]}} & \makecell{0.948\\ {[0.947, 0.948]}} & 0.214 & 0.243 & \makecell{\textbf{0.903}\\ {[0.902, 0.904]}} & \makecell{0.810\\ {[0.808, 0.811]}} & 0.307 & 0.401 \\
\midrule
I+II+V1 & PTB-XL & E+LF & \makecell{0.989\\ {[0.989, 0.989]}} & \makecell{0.979\\ {[0.979, 0.980]}} & 0.145 & 0.156 & \makecell{\textbf{0.964}\\ {[0.962, 0.965]}} & \makecell{0.930\\ {[0.927, 0.932]}} & 0.203 & 0.280 \\
 & Georgia & E+LF & \makecell{0.983\\ {[0.983, 0.983]}} & \makecell{0.978\\ {[0.977, 0.978]}} & 0.188 & 0.181 & \makecell{\textbf{0.924}\\ {[0.923, 0.926]}} & \makecell{0.900\\ {[0.898, 0.901]}} & 0.279 & 0.322 \\
 & China & E+LF & \makecell{0.983\\ {[0.983, 0.984]}} & \makecell{0.973\\ {[0.972, 0.973]}} & 0.173 & 0.187 & \makecell{\textbf{0.943}\\ {[0.943, 0.944]}} & \makecell{0.904\\ {[0.903, 0.905]}} & 0.250 & 0.333 \\
\midrule
I+II+V2 & PTB-XL & E+LF & \makecell{0.991\\ {[0.990, 0.991]}} & \makecell{0.985\\ {[0.984, 0.985]}} & 0.143 & 0.139 & \makecell{\textbf{0.971}\\ {[0.970, 0.972]}} & \makecell{0.950\\ {[0.949, 0.952]}} & 0.193 & 0.248 \\
 & Georgia & E+LF & \makecell{0.987\\ {[0.986, 0.987]}} & \makecell{0.984\\ {[0.984, 0.985]}} & 0.182 & 0.167 & \makecell{\textbf{0.939}\\ {[0.938, 0.940]}} & \makecell{0.927\\ {[0.926, 0.928]}} & 0.262 & 0.297 \\
 & China & E+LF & \makecell{0.986\\ {[0.985, 0.986]}} & \makecell{0.979\\ {[0.979, 0.979]}} & 0.175 & 0.167 & \makecell{\textbf{0.949}\\ {[0.949, 0.950]}} & \makecell{0.926\\ {[0.926, 0.927]}} & 0.247 & 0.297 \\
\midrule
I+II+V1+V4 & PTB-XL & E+LF & \makecell{0.995\\ {[0.995, 0.995]}} & \makecell{0.999\\ {[0.998, 0.999]}} & 0.126 & 0.112 & \makecell{\textbf{0.979}\\ {[0.978, 0.979]}} & \makecell{0.962\\ {[0.961, 0.964]}} & 0.174 & 0.222 \\
 & Georgia & E+LF & \makecell{0.994\\ {[0.994, 0.994]}} & \makecell{0.999\\ {[0.999, 0.999]}} & 0.151 & 0.133 & \makecell{\textbf{0.962}\\ {[0.961, 0.963]}} & \makecell{0.947\\ {[0.946, 0.947]}} & 0.224 & 0.261 \\
 & China & E+LF & \makecell{0.993\\ {[0.993, 0.993]}} & \makecell{0.999\\ {[0.999, 0.999]}} & 0.151 & 0.134 & \makecell{\textbf{0.965}\\ {[0.965, 0.965]}} & \makecell{0.947\\ {[0.947, 0.948]}} & 0.219 & 0.263 \\
\bottomrule
\end{tabular}
\end{table}

For configurations containing I and II, the four remaining limb leads are exactly derivable from the inputs. Einthoven + linear filter reproduces them exactly (per-lead PCC $>0.9999$), and BiMa-SR learned them to a per-lead PCC of 0.996--0.997 from I+II. Aggregates over all predicted leads therefore favour the comparator once few chest leads remain to be predicted (I+II+V1+V4 on PTB-XL: 0.995 for BiMa-SR against 0.999), and the fair comparison is on the predicted chest leads, which we use as the primary metric for these configurations. On the chest leads, BiMa-SR reached 0.939 against 0.849 from I+II, 0.964 against 0.930 from I+II+V1, 0.971 against 0.950 from I+II+V2 and 0.979 against 0.962 from I+II+V1+V4 on PTB-XL (Fig.~\ref{fig:2}a, Table~\ref{tab:1}); the chest-lead MAE followed the same pattern, for example 0.174 against 0.222 from I+II+V1+V4 (Fig.~\ref{fig:2}c).

Across the six configurations and three cohorts, the chest-lead PCC of BiMa-SR exceeded that of the comparator in all 18 cases, by 0.012 to 0.136. The margin decreased as measured chest leads were added: on PTB-XL it was 0.136 and 0.132 for leads I and II alone, 0.090 for I+II, 0.034 for I+II+V1, 0.021 for I+II+V2 and 0.016 for I+II+V1+V4. The paired per-record analysis agreed: BiMa-SR had the higher per-record chest-lead PCC in 92.7\% of PTB-XL test records from lead I and in 77.2\% from I+II+V1+V4, and in 60.2\% to 92.7\% of records across all 18 configuration--cohort pairs (Wilcoxon signed-rank $P<10^{-80}$ in every case). On the external cohorts, the chest-lead PCC of BiMa-SR ranged from 0.849 (lead I, Georgia) to 0.965 (I+II+V1+V4, China); per-lead values are in Supplementary Fig.~S2 and all groups with confidence intervals in Supplementary Table~S1.

\subsection{Single-lead reconstruction: regression versus diffusion}

From lead II, the diffusion model DiBiMa-SR, a denoising diffusion model with $T=20$ steps sharing the BiMa-SR backbone, was evaluated with a single network evaluation (one-step) and with a six-step DDIM sampler (DDIM-6), both starting from pure Gaussian noise. Over all 11 predicted leads, BiMa-SR reached a pooled median PCC of 0.914 on PTB-XL, DiBiMa-SR one-step 0.903 [0.900, 0.907], DiBiMa-SR DDIM-6 0.862 [0.859, 0.865] and the linear filter 0.784 (Fig.~\ref{fig:2}d). The same order held on Georgia (0.871, 0.862, 0.819, 0.772) and China (0.880, 0.866, 0.827, 0.789). The per-record PCC difference between BiMa-SR and DiBiMa-SR one-step was +0.017 [+0.015, +0.020] on PTB-XL (BiMa-SR higher in 62.8\% of records; Wilcoxon $P=3.2\times10^{-48}$), +0.012 [+0.011, +0.014] on Georgia (57.8\%; $P=8.5\times10^{-71}$) and +0.024 [+0.024, +0.025] on China (65.9\%; $P<10^{-300}$).

Within DiBiMa-SR, one-step sampling was more accurate than DDIM-6 in every predicted lead and cohort (Supplementary Fig.~S3): the per-record PCC difference, one-step minus DDIM-6, was +0.039 [+0.038, +0.040] on PTB-XL, +0.034 [+0.034, +0.035] on Georgia and +0.033 [+0.033, +0.033] on China, with one-step better in 91.2\%, 86.9\% and 87.6\% of records, and the MAE was lower with one-step sampling in 99.8\%, 99.3\% and 99.7\% of records. Because the one-step prediction is the network's estimate of the conditional mean of the target given the input, whereas multi-step sampling draws a sample from the learned conditional distribution, a pointwise metric such as PCC or MAE is expected to favour the former; additional sampling steps did not improve pointwise accuracy here.

\subsection{Qualitative reconstructions}

\begin{figure}[tbp]
\centering
\includegraphics[width=\textwidth]{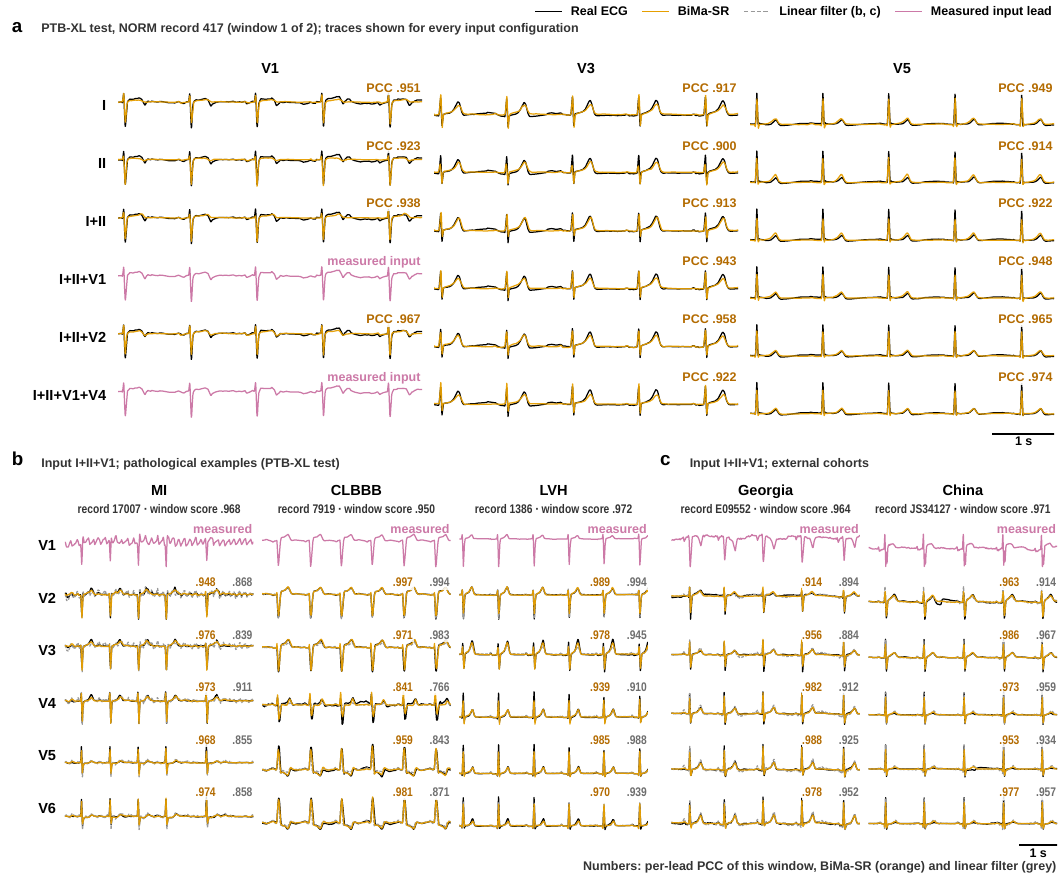}
\caption{\textbf{Qualitative reconstructions.}
Every example is selected by a fixed rule, not by eye: within a group, windows are ranked by a per-window score (BiMa-SR PCC averaged over the scored leads of that window) and the window at the 90th percentile (nearest rank) is shown, a good but typical case; the same groups at the 10th percentile are in Supplementary Fig.~S5. Traces: 5-s windows at 500\,Hz in z units (each lead scaled separately; bar, 1\,s). Black, real ECG; orange, BiMa-SR; dashed grey, linear filter; pink, measured input lead. Numbers: per-lead PCC of the shown window (orange, BiMa-SR; grey, linear filter).
\textbf{a},~PTB-XL test, NORM records; score over the 11 predicted leads of the lead II model (1,926 windows / 963 records); record 417, window 1 of 2, score 0.936. The same window reconstructed by BiMa-SR from each of the six configurations (rows); V1, V3 and V5 shown.
\textbf{b},~Input I+II+V1, PTB-XL test, one window per diagnostic label (a record may carry more than one label); score over the predicted chest leads V2--V6. MI: record 17007, window 1 of 2, score 0.968 (1,100 windows / 550 records); CLBBB: record 7919, window 1 of 2, score 0.950 (108 / 54); LVH: record 1386, window 1 of 2, score 0.972 (428 / 214). The comparator for I+II+V1 is Einthoven + linear filter, whose chest leads are the linear-filter chest leads shown.
\textbf{c},~Input I+II+V1, external cohorts (all windows), same score. Georgia: record E09552, window 1 of 2, score 0.964 (20,547 windows / 10,275 records); China: record JS34127, window 1 of 2, score 0.971 (87,105 / 43,559).}
\label{fig:3}
\end{figure}

To show typical rather than selected behaviour, examples were chosen by a fixed rule: the window at the 90th percentile of the BiMa-SR per-window score within a group (Fig.~\ref{fig:3}) and, for the same groups, the window at the 10th percentile (Supplementary Fig.~S5). At the 90th percentile, the reconstructed chest leads closely overlaid the real ECG for a PTB-XL NORM record from every input configuration (score 0.936; Fig.~\ref{fig:3}a), for records labelled MI, CLBBB and LVH from I+II+V1 (scores 0.968, 0.950 and 0.972; Fig.~\ref{fig:3}b) and for the external cohorts (Georgia 0.964, China 0.971; Fig.~\ref{fig:3}c); BiMa-SR had the higher per-lead PCC in 22 of the 25 shown predicted leads that also carry a linear-filter trace. At the 10th percentile (scores 0.620--0.768), reconstructions missed morphology in individual chest leads, and the linear filter was closer in 6 of 25 shown leads (for example LVH, V3: 0.244 against 0.450). Low-percentile windows included recordings with noisy or atypical leads that neither method followed (Georgia, V3: PCC $-0.362$ for BiMa-SR and $-0.159$ for the linear filter).

\subsection{Diagnostic readability}

\begin{figure}[p]
\centering
\includegraphics[width=\textwidth]{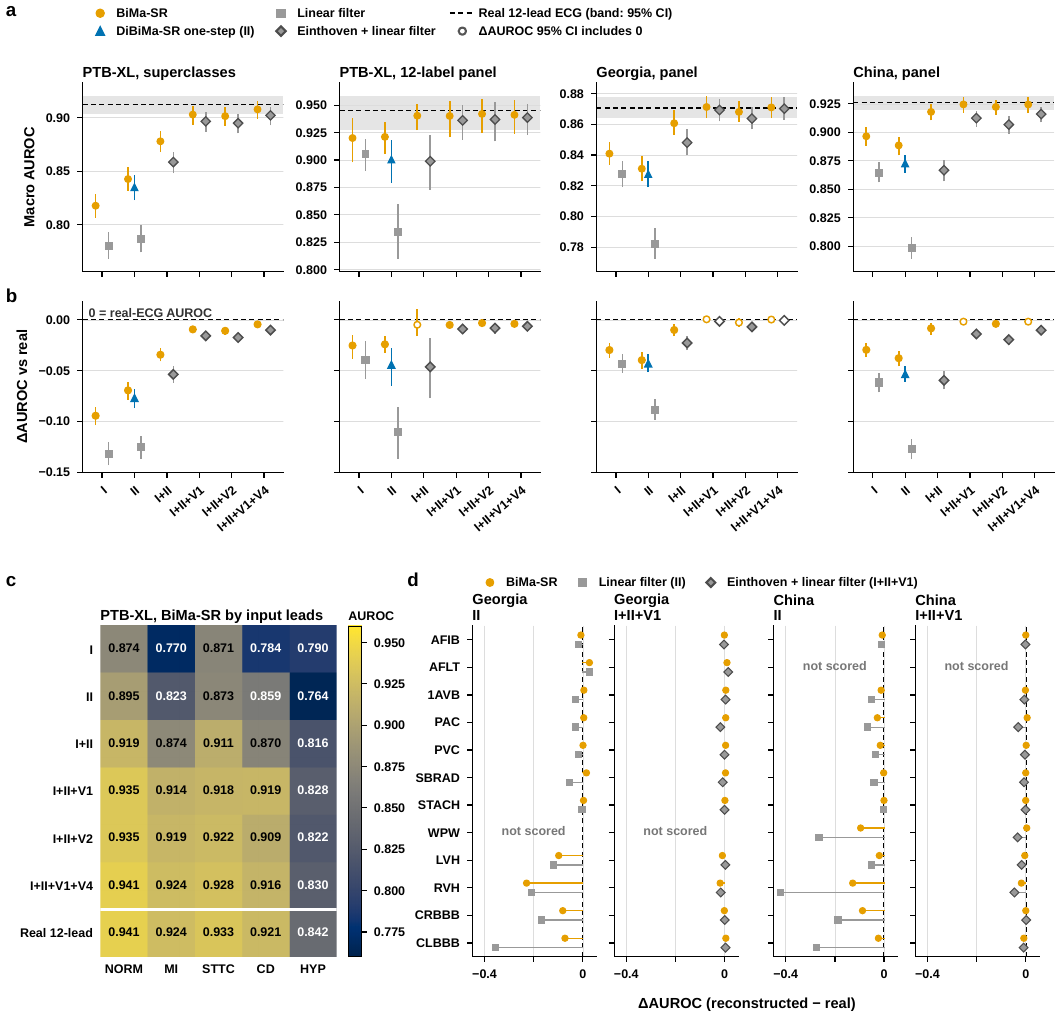}
\caption{\textbf{Diagnostic readability of reconstructed ECGs by a classifier trained on real ECGs only.}
One 12-lead 1D-ResNet (3.36\,M parameters), trained only on real PTB-XL ECGs (folds 1--8), is applied unchanged to real 12-lead ECGs and to 12-lead ECGs reconstructed from each configuration (measured leads re-inserted). Scores are averaged over the two 5-s windows of a record; AUROCs are record level. Cohorts: PTB-XL test ($n=2{,}198$ records; 2,158 with a superclass label), Georgia ($n=10{,}275$) and China ($n=43{,}559$); records with an unknown label are excluded from that label. 95\% CIs: record bootstrap, $B=1{,}000$, identical resamples for every input type. Baseline: linear filter for I and II; Einthoven + linear filter for configurations containing I and II.
\textbf{a},~Macro AUROC (95\% CI) for BiMa-SR (orange circles), the baseline (grey squares: linear filter; diamonds: Einthoven + linear filter) and DiBiMa-SR one-step (blue triangle, II only), on the PTB-XL superclasses (NORM, MI, STTC, CD, HYP) and on the 12-label panel (AFIB, AFLT, 1AVB, PAC, PVC, SBRAD, STACH, WPW, LVH, RVH, CRBBB, CLBBB) of PTB-XL, Georgia and China. Dashed line and band: real 12-lead ECG and its 95\% CI. Labels with fewer than 10 positive records are skipped (PTB-XL AFLT, 7; PTB-XL WPW, 8; Georgia WPW, 2); China AFLT is not scored. Example: PTB-XL superclasses, real 0.912 (0.903--0.920); from lead II, BiMa-SR 0.843 (0.832--0.854), linear filter 0.787 (0.774--0.800); from I+II+V1+V4, BiMa-SR 0.908 (0.899--0.916).
\textbf{b},~Difference in macro AUROC between reconstructed and real ECGs (mean over paired bootstrap resamples, 95\% percentile CI). Filled markers: CI excludes 0; open markers: CI includes 0. BiMa-SR differences whose CI includes 0: I+II, PTB-XL panel; I+II+V1, I+II+V2 and I+II+V1+V4, Georgia panel; I+II+V1 and I+II+V1+V4, China panel. Example, China panel from lead II: BiMa-SR $-0.038$ ($-0.046$ to $-0.030$), linear filter $-0.128$ ($-0.137$ to $-0.117$).
\textbf{c},~Per-superclass AUROC on PTB-XL ECGs reconstructed by BiMa-SR (one row per configuration) and on real ECGs (bottom row), $n=2{,}158$ records.
\textbf{d},~Per-label AUROC difference (reconstructed $-$ real; point estimates) on the external cohorts for II and I+II+V1. Orange circles: BiMa-SR; grey square: linear filter (II); grey diamond: Einthoven + linear filter (I+II+V1). Not scored: Georgia WPW and China AFLT. Per-label AUROCs with 95\% CIs are in Supplementary Fig.~S4.}
\label{fig:4}
\end{figure}

A 12-lead 1D-ResNet trained only on real PTB-XL ECGs reached a record-level macro AUROC of 0.912 [0.903, 0.920] on the five superclasses of the real PTB-XL test ECGs, and 0.945, 0.871 and 0.926 on the 12-label panel of PTB-XL, Georgia and China (Fig.~\ref{fig:4}a, Table~\ref{tab:2}). Applied unchanged to reconstructions, its superclass macro AUROC was 0.843 [0.832, 0.854] from lead II and 0.818 from lead I (linear filter 0.787 and 0.780), 0.878 from I+II (Einthoven + linear filter 0.859), and 0.903, 0.901 and 0.908 from I+II+V1, I+II+V2 and I+II+V1+V4. The paired differences from real ECGs were $-0.069$ [$-0.079$, $-0.061$] from lead II and $-0.009$, $-0.011$ and $-0.004$ with three or four measured leads (Fig.~\ref{fig:4}b). The point estimate for BiMa-SR reconstructions was higher than for the linear comparator in every configuration and cohort; the difference was largest from lead II (superclasses +0.055 [+0.048, +0.064]; China panel +0.090 [+0.080, +0.099]) and was small with three or four measured leads (+0.001 to +0.016; Table~\ref{tab:2}). DiBiMa-SR one-step reconstructions from lead II scored 0.835 on the superclasses, 0.008 [0.004, 0.011] below BiMa-SR.

\begin{table}[tbp]
\centering
\caption{\textbf{Diagnostic readability: macro AUROC of a classifier trained on real ECGs only, applied to real and reconstructed ECGs.} Record-level macro AUROC with 95\% record-bootstrap CI ($B=1{,}000$; identical resamples for every input). $\Delta$: mean of the paired bootstrap differences with 95\% percentile CI (this mean can differ by 0.001 from the difference of the rounded point estimates). Comparator: LF, linear filter (I, II); E+LF, Einthoven + linear filter (configurations containing I and II). Labels with fewer than 10 positive records are not scored (PTB-XL AFLT and WPW; Georgia WPW); AFLT is never scored in China. DiBiMa-SR one-step from lead II: 0.835 [0.823, 0.847] (PTB-XL superclasses), 0.901 [0.879, 0.919] (PTB-XL panel), 0.828 [0.819, 0.836] (Georgia) and 0.873 [0.865, 0.880] (China); BiMa-SR minus DiBiMa-SR: +0.008 [+0.004, +0.011], +0.020 [+0.007, +0.040], +0.003 [$-$0.002, +0.009] and +0.016 [+0.011, +0.021]. Per-label values: Supplementary Tables~S2a and S2b.}
\label{tab:2}
\footnotesize
\setlength{\tabcolsep}{3.5pt}
\begin{tabular}{@{}lcccc@{}}
\toprule
Input & BiMa-SR & Comparator & $\Delta$ BiMa-SR $-$ real & $\Delta$ BiMa-SR $-$ comparator \\
 & AUROC [95\% CI] & AUROC [95\% CI] & [95\% CI] & [95\% CI] \\
\midrule
\multicolumn{5}{@{}l}{\textit{PTB-XL test, 5 diagnostic superclasses}; real 12-lead ECG 0.912 [0.903, 0.920]} \\
\quad I & 0.818 [0.806, 0.828] & 0.780 [0.768, 0.793] & $-$0.094 [$-$0.104, $-$0.085] & +0.038 [+0.027, +0.048] \\
\quad II & 0.843 [0.832, 0.854] & 0.787 [0.774, 0.800] & $-$0.069 [$-$0.079, $-$0.061] & +0.055 [+0.048, +0.064] \\
\quad I+II & 0.878 [0.868, 0.887] & 0.859 [0.848, 0.868] & $-$0.034 [$-$0.041, $-$0.028] & +0.019 [+0.012, +0.027] \\
\quad I+II+V1 & 0.903 [0.893, 0.911] & 0.897 [0.887, 0.905] & $-$0.009 [$-$0.013, $-$0.006] & +0.006 [+0.003, +0.010] \\
\quad I+II+V2 & 0.901 [0.892, 0.910] & 0.895 [0.885, 0.903] & $-$0.011 [$-$0.015, $-$0.007] & +0.007 [+0.003, +0.010] \\
\quad I+II+V1+V4 & 0.908 [0.899, 0.916] & 0.902 [0.893, 0.910] & $-$0.004 [$-$0.007, $-$0.002] & +0.006 [+0.002, +0.009] \\
\addlinespace[3pt]
\multicolumn{5}{@{}l}{\textit{PTB-XL test, 12-label panel}; real 12-lead ECG 0.945 [0.927, 0.959]} \\
\quad I & 0.920 [0.898, 0.939] & 0.906 [0.890, 0.919] & $-$0.025 [$-$0.038, $-$0.015] & +0.015 [$-$0.008, +0.036] \\
\quad II & 0.921 [0.905, 0.935] & 0.835 [0.810, 0.860] & $-$0.024 [$-$0.032, $-$0.016] & +0.086 [+0.059, +0.115] \\
\quad I+II & 0.941 [0.927, 0.952] & 0.899 [0.873, 0.923] & $-$0.005 [$-$0.016, +0.011] & +0.041 [+0.017, +0.069] \\
\quad I+II+V1 & 0.940 [0.921, 0.954] & 0.936 [0.918, 0.950] & $-$0.005 [$-$0.008, $-$0.002] & +0.004 [$-$0.000, +0.009] \\
\quad I+II+V2 & 0.942 [0.925, 0.956] & 0.937 [0.917, 0.953] & $-$0.003 [$-$0.007, $-$0.000] & +0.005 [+0.000, +0.010] \\
\quad I+II+V1+V4 & 0.941 [0.924, 0.955] & 0.939 [0.923, 0.952] & $-$0.004 [$-$0.006, $-$0.002] & +0.002 [$-$0.002, +0.006] \\
\addlinespace[3pt]
\multicolumn{5}{@{}l}{\textit{Georgia (external), 12-label panel}; real 12-lead ECG 0.871 [0.864, 0.878]} \\
\quad I & 0.841 [0.834, 0.848] & 0.828 [0.819, 0.836] & $-$0.030 [$-$0.037, $-$0.023] & +0.013 [+0.002, +0.024] \\
\quad II & 0.831 [0.823, 0.840] & 0.782 [0.773, 0.792] & $-$0.040 [$-$0.048, $-$0.031] & +0.049 [+0.040, +0.058] \\
\quad I+II & 0.861 [0.853, 0.869] & 0.848 [0.840, 0.857] & $-$0.010 [$-$0.016, $-$0.004] & +0.013 [+0.005, +0.021] \\
\quad I+II+V1 & 0.871 [0.864, 0.879] & 0.869 [0.862, 0.877] & +0.001 [$-$0.003, +0.004] & +0.002 [$-$0.002, +0.006] \\
\quad I+II+V2 & 0.868 [0.862, 0.875] & 0.864 [0.857, 0.871] & $-$0.003 [$-$0.007, +0.001] & +0.004 [+0.000, +0.008] \\
\quad I+II+V1+V4 & 0.871 [0.864, 0.878] & 0.870 [0.863, 0.878] & +0.000 [$-$0.003, +0.003] & +0.001 [$-$0.003, +0.004] \\
\addlinespace[3pt]
\multicolumn{5}{@{}l}{\textit{China (external), 12-label panel}; real 12-lead ECG 0.926 [0.919, 0.932]} \\
\quad I & 0.897 [0.888, 0.904] & 0.865 [0.856, 0.873] & $-$0.030 [$-$0.037, $-$0.023] & +0.032 [+0.022, +0.042] \\
\quad II & 0.888 [0.880, 0.896] & 0.799 [0.789, 0.808] & $-$0.038 [$-$0.046, $-$0.030] & +0.090 [+0.080, +0.099] \\
\quad I+II & 0.918 [0.911, 0.925] & 0.867 [0.857, 0.875] & $-$0.008 [$-$0.015, $-$0.003] & +0.051 [+0.042, +0.060] \\
\quad I+II+V1 & 0.924 [0.917, 0.931] & 0.912 [0.905, 0.919] & $-$0.002 [$-$0.006, +0.002] & +0.012 [+0.008, +0.017] \\
\quad I+II+V2 & 0.922 [0.915, 0.929] & 0.907 [0.899, 0.914] & $-$0.004 [$-$0.007, $-$0.000] & +0.016 [+0.011, +0.020] \\
\quad I+II+V1+V4 & 0.924 [0.917, 0.931] & 0.916 [0.909, 0.923] & $-$0.002 [$-$0.005, +0.001] & +0.008 [+0.005, +0.012] \\
\bottomrule
\end{tabular}
\end{table}

On the external panels, the gap to real ECGs from lead II was $-0.040$ [$-0.048$, $-0.031$] on Georgia and $-0.038$ [$-0.046$, $-0.030$] on China, against $-0.089$ and $-0.128$ for the linear filter. With three or four measured leads, the BiMa-SR differences from real ECGs on the external panels were between $-0.004$ and +0.001, and the 95\% CI included zero for I+II+V1, I+II+V2 and I+II+V1+V4 on Georgia and for I+II+V1 and I+II+V1+V4 on China (Fig.~\ref{fig:4}b). Per superclass (Fig.~\ref{fig:4}c), MI depended most on the measured chest leads (AUROC 0.823 from lead II, 0.914 from I+II+V1, 0.924 on real ECGs), and HYP was the hardest class in every row except lead I, where MI was lowest (HYP: 0.764 from lead II, 0.830 from I+II+V1+V4, 0.842 on real ECGs; MI from lead I: 0.770). Per-label differences on the external cohorts (Fig.~\ref{fig:4}d) and per-label AUROCs with confidence intervals (Supplementary Fig.~S4, Supplementary Tables~S2a and S2b) show the same pattern.

\subsection{External generalisation, diagnostic subgroups and efficiency}

\begin{figure}[tbp]
\centering
\includegraphics[width=\textwidth]{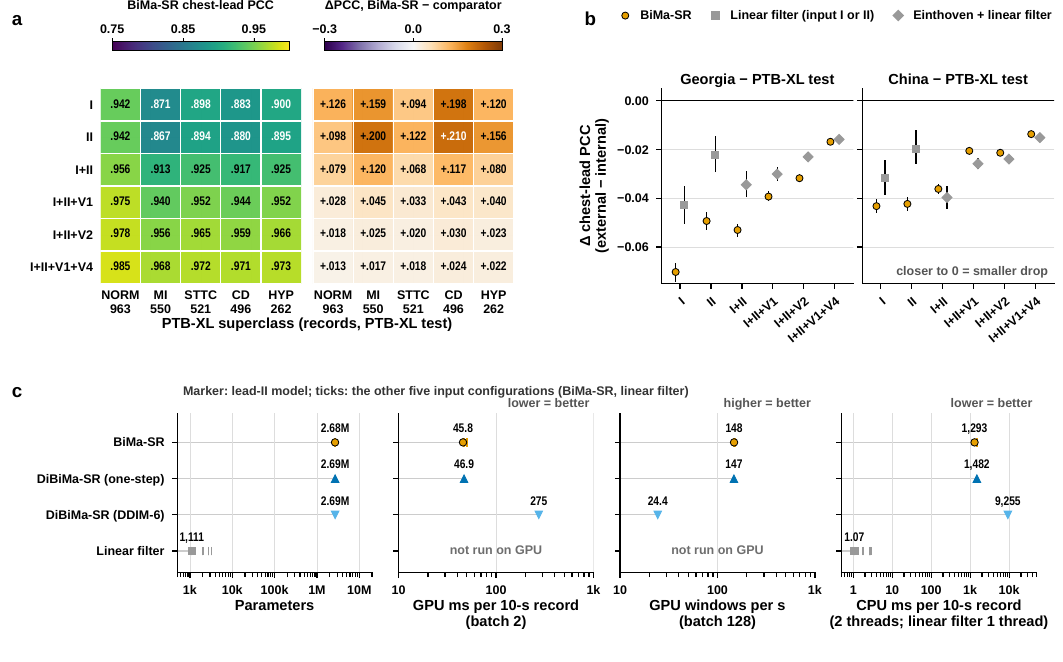}
\caption{\textbf{Generalisation, diagnostic subgroups and efficiency.}
Accuracy is the chest-lead pooled median PCC (z-scored 5-s windows), used for every configuration for consistency with Fig.~\ref{fig:2}a. Comparator: linear filter for I or II alone; Einthoven + linear filter otherwise (identical chest leads).
\textbf{a},~PTB-XL test, per diagnostic superclass (NORM, normal; MI, myocardial infarction; STTC, ST/T change; CD, conduction disturbance; HYP, hypertrophy); a record contributes to every superclass it carries; the number under each label is the number of records. Left, BiMa-SR; right, BiMa-SR minus comparator. BiMa-SR is higher in all 30 cells (smallest difference +0.013, largest +0.210).
\textbf{b},~External-cohort change: pooled median PCC on Georgia (10,275 records) or China (43,559 records) minus PTB-XL test (2,198 records). 95\% CIs: percentile intervals of the difference between independent record bootstraps of the two cohorts ($B=1{,}000$ each). Both methods lose accuracy externally. The BiMa-SR drop is larger than the comparator's (non-overlapping CIs) for I, II, I+II, I+II+V1 and I+II+V2 on Georgia and for I and II on China, and smaller for I+II+V1 on China; BiMa-SR remains above the comparator in absolute chest-lead PCC in every configuration and cohort (Fig.~\ref{fig:2}a). Examples: input I, Georgia $-0.070$ [$-0.074$, $-0.067$] for BiMa-SR versus $-0.043$ [$-0.051$, $-0.035$] for the linear filter; I+II+V1+V4, China $-0.014$ [$-0.015$, $-0.013$] versus $-0.015$ [$-0.017$, $-0.014$].
\textbf{c},~Efficiency. Markers: lead-II models; vertical ticks: the other five configurations (BiMa-SR 2,675,148--2,676,108 parameters; linear filter 1,111--3,232 coefficients $=$ 101 taps $\times$ input leads $\times$ predicted leads). GPU latency per 10-s record (batch of its two 5-s windows); throughput in windows per second at batch 128; CPU latency per 10-s record with 2 threads (linear filter: 1 thread). The linear filter was not timed on GPU. Lead II: BiMa-SR 45.8\,ms (GPU) and 1,293\,ms (CPU); DiBiMa-SR one-step 46.9\,ms and 1,482\,ms; DDIM-6 (6 network evaluations) 274.5\,ms and 9,255\,ms; linear filter 1.07\,ms (CPU, 1 thread). Hardware and timing protocol in Methods.}
\label{fig:5}
\end{figure}

Within PTB-XL, BiMa-SR was more accurate than the comparator in every diagnostic superclass and configuration (30 of 30 cells; differences +0.013 to +0.210; Fig.~\ref{fig:5}a). The largest gain was in conduction disturbance from lead II (0.880 against 0.670), and the smallest in NORM from I+II+V1+V4 (0.985 against 0.972).

Both methods lost accuracy on the external cohorts (Fig.~\ref{fig:5}b). The chest-lead PCC drop from PTB-XL to Georgia was larger for BiMa-SR than for the comparator in five configurations, for example $-0.070$ [$-0.074$, $-0.067$] against $-0.043$ [$-0.051$, $-0.035$] from lead I, and the drops converged as measured chest leads were added ($-0.017$ against $-0.016$ from I+II+V1+V4). On China, the BiMa-SR drop was larger from I and II, smaller from I+II+V1, and indistinguishable for the other configurations (I+II+V1+V4: $-0.014$ [$-0.015$, $-0.013$] against $-0.015$ [$-0.017$, $-0.014$]). In absolute terms, BiMa-SR remained above the comparator in all 12 external configuration--cohort pairs (Table~\ref{tab:1}).

BiMa-SR is small and fast (Fig.~\ref{fig:5}c; Supplementary Table~S3). From lead II it has 2,675,148 parameters and reconstructs a 10-s record (two 5-s windows) in 45.8\,ms on a GPU, with a throughput of 148 windows per second at batch 128, and in 1,293\,ms on two CPU threads. The other configurations took 46.2 to 50.3\,ms per record on the GPU. DiBiMa-SR one-step had a similar cost (46.9\,ms GPU, 1,482\,ms CPU), whereas DDIM-6 required six network evaluations (274.5\,ms GPU, 9,255\,ms CPU). The linear filter, with 1,111 coefficients for one input lead, took 1.07\,ms per record on one CPU thread.

\section{Discussion}

A single bidirectional state-space backbone, trained with one recipe, reconstructed the 12-lead ECG from six reduced lead sets more accurately than least-squares linear filters on an internal test fold and on two external cohorts. The models receive only the measured leads at inference, the checkpoint is chosen on a fixed PTB-XL validation subset, and the external cohorts play no part in training or model selection, so the external results measure transfer to new devices, populations and labelling practices. Because the four derived limb leads are exactly reconstructable from I and II, we reported the predicted chest leads as the primary metric for input sets containing I and II; on all predicted leads, the exact Einthoven + linear filter comparator is in fact slightly ahead for I+II+V1+V4 (0.999 against 0.995 on PTB-XL), a reminder that all-lead averages can hide where reconstruction error lies.

The advantage of the learned model was largest where the linear problem is hardest. From a single limb lead, the linear filter cannot recover leads nearly orthogonal to the input (from lead I, PCC 0.390 in aVF and 0.557 in V2), whereas BiMa-SR reached 0.713 and 0.892. As measured chest leads were added, the margin over linear baselines diminished: on PTB-XL it fell from 0.136 (lead I) to 0.090 (I+II), 0.034 (I+II+V1) and 0.016 (I+II+V1+V4), and the classifier-level advantage over the comparator with three or four measured leads was between +0.001 and +0.016 in macro AUROC. This agrees with classical work in which general linear transforms performed well when only one or two precordial leads were missing~\cite{nelwan2004reconstruction,horacek2008designing,drew2002comparison}. With four measured leads, a well-fitted linear filter is a strong, transparent and nearly free alternative (1.07\,ms per record on one CPU thread), and a learned model adds a small, consistent improvement rather than a qualitative change.

From lead II, regression was more accurate than diffusion with the same backbone, and one-step diffusion was more accurate than six-step sampling. Under a zero-terminal-SNR schedule with $x_0$-prediction, the one-step output from pure noise is an estimate of the conditional mean of the 12-lead ECG given the measured lead, which is what pointwise metrics such as PCC and MAE reward; multi-step sampling instead draws a sample, which adds variability that these metrics penalise. Diffusion may still be useful where a distribution of plausible reconstructions is wanted, for example to quantify uncertainty or to counter the amplitude shrinkage reported for regression-type reconstructions~\cite{presacan2025evaluating}, but we did not evaluate those uses here.

The clinical scenarios differ in how much a reconstruction can offer. For a smartwatch lead I, superclass AUROC on reconstructions was 0.818 against 0.912 on real ECGs, so a large part of the 12-lead diagnostic signal is not recoverable from that lead. For five-electrode telemetry (I+II+V1), the classifier read reconstructions with a superclass AUROC of 0.903 and, on the external panels, within 0.002 of real ECGs, which suggests that a 12-lead tool could be run on telemetry recordings with little loss for the labels studied here. Reconstruction cannot, however, add information absent from the measured leads: every reconstructed lead is a function of the inputs (and, for diffusion, of random noise), so any diagnostic information in the reconstruction was already present in the measured leads. High classifier readability therefore means that reconstructed ECGs are usable by existing 12-lead tools, not that new diagnostic information has been created; a reconstruction that looks plausible can still be wrong for an individual patient, as the low-percentile examples show.

This study has limitations. (i) Evaluation was on z-scored leads: each lead of each window was standardised, PCC is invariant to amplitude, and MAE is in z units, so absolute amplitudes, inter-lead amplitude ratios and amplitude-based criteria (for example voltage criteria for hypertrophy) were not assessed; the classifier also received z-scored input, which mainly penalises HYP. The models do receive inputs with preserved inter-lead ratios (common scale $s$), and the stored per-lead statistics allow amplitude-level evaluation in future work. (ii) Inputs were standard-electrode leads extracted from 12-lead recordings; modified Holter leads, EASI and Frank electrode positions cannot be simulated from these data, and real wearables differ in electrode placement, contact and noise. (iii) The external cohorts differ from PTB-XL in devices, populations and label sources, and their labels are annotation-reference labels mapped from SNOMED-CT codes. In China, code 164909002, which we mapped to CLBBB following the PhysioNet/CinC 2021 equivalences, also covers left anterior and posterior fascicular blocks in the database dictionary, and the AFLT labels are unusable (all 7,959 AFLT-positive records carry a code that the canonical mapping treats as ambiguous for AFIB), so AFLT was not scored. (iv) Each model was trained with a single seed, DiBiMa-SR only for lead II, and the classifier is a single model; other classifiers may read reconstructions differently. (v) There was no prospective or clinical validation and no reading by cardiologists. BiMa-ECGSR, including its online demonstration, is a research prototype and is not intended for clinical use.

\section{Methods}

\subsection{Datasets and labels}

We used three public 12-lead databases from PhysioNet~\cite{goldberger2000physionet}: PTB-XL v1.0.3 (\texttt{records500}; 21,799 records from Germany)~\cite{wagner2020ptbxl}, the Georgia 12-lead ECG database (10,344 records from the USA) distributed with the PhysioNet/Computing in Cardiology Challenges 2020 and 2021~\cite{perezalday2020classification,reyna2021will}, and the Chapman-Shaoxing/Ningbo database (\texttt{ecg-arrhythmia} v1.0.0; 45,152 records from China)~\cite{zheng2020chapman,zheng2020optimal,zheng2022largescale}. No records were subsampled. PTB-XL was split by its recommended stratified, patient-disjoint folds (folds 1--8 training, 9 validation, 10 test)~\cite{wagner2020ptbxl,strodthoff2021deep}; Georgia and China were used only as external test sets.

PTB-XL diagnostic superclasses (NORM, MI, STTC, CD, HYP) were derived as multi-hot labels from all SCP codes with \texttt{diagnostic}$=1$, regardless of likelihood. A shared 12-label rhythm and morphology panel (AFIB, AFLT, 1AVB, PAC, PVC, SBRAD, STACH, WPW, LVH, RVH, CRBBB, CLBBB) was defined by SCP-code presence in PTB-XL and by a fixed SNOMED-CT mapping in the external cohorts (for example PVC $=$ 17338001 or 427172004; CRBBB $=$ 713427006 or 59118001; CLBBB $=$ 733534002 or 164909002, following the PhysioNet/CinC 2021 scoring equivalences~\cite{reyna2021will,reyna2022issues}; incomplete RBBB, 713426002, is not CRBBB). External labels are annotation-reference labels: a label is unknown when the record has no diagnosis code or carries a code that is ambiguous for that target (195080001 or 164890007 for AFIB, 195080001 for AFLT, 63593006 for PAC), and negative otherwise. In China, all 7,959 AFLT-positive records carry SNOMED 164890007, the code treated as ambiguous for AFIB, the dictionary assigns 164909002 both to LBBB and to left anterior/posterior fascicular block, and 34 records are positive for both CRBBB and CLBBB. Label counts are in Supplementary Table~S4.

\subsection{Preprocessing}

All ECGs were processed by one pipeline. Records were read with \texttt{wfdb} in physical units (mV) and the leads reordered by header name to I, II, III, aVR, aVL, aVF, V1--V6. All readable records were natively sampled at 500\,Hz, so no resampling was needed; the first 10\,s (5,000 samples) were kept and shorter records excluded. Each record was band-pass filtered with a zero-phase third-order Butterworth filter (0.5--40\,Hz, second-order sections, \texttt{scipy.signal.sosfiltfilt}), split into two non-overlapping 5-s windows (2,500 samples), and each lead of each window was z-scored with its own mean and standard deviation; the per-lead mean and standard deviation (mV) were stored so that absolute amplitudes can be restored as $x_\mathrm{mV}=z\,\sigma+\mu$. A window was excluded if it contained a non-finite value or a lead with a standard deviation below 0.005\,mV (flat lead). This gave 34,833 / 4,366 / 4,396 windows from 17,417 / 2,183 / 2,198 PTB-XL training/validation/test records, 20,547 windows from 10,275 Georgia records and 87,105 windows from 43,559 China records. PTB-XL training record 10565, which contains ADC-saturated samples, was excluded from reconstruction training and from the linear-filter fit. For model selection, a fixed subset of 1,024 PTB-XL validation windows was drawn with \texttt{numpy.random.default\_rng(0)}.

\subsection{Input normalisation}

For an input set of $k$ leads, the measured leads of a window were destandardised to mV, mean-centred and divided by one common scale
\begin{equation}
s=\Big(\tfrac{1}{k}\textstyle\sum_{j=1}^{k}\sigma_j^2\Big)^{1/2},
\end{equation}
where $\sigma_j$ is the standard deviation of input lead $j$ in the window. A common scale preserves the amplitude ratios between the measured leads, which carry information about the electrical axis, and equals the per-lead z-score when $k=1$. The target was the stored per-lead z-score of all 12 leads.

\subsection{Architecture}

BiMa-SR and DiBiMa-SR share one backbone (Fig.~\ref{fig:1}c). A convolutional encoder (Conv1d, kernel 5, 64 channels, batch normalisation, SiLU, dropout 0.3) maps the input channels to a 64-dimensional sequence at the full 500-Hz resolution. A bidirectional Mamba stack~\cite{gu2023mamba,zhu2024vision,schiff2024caduceus} of 8 layers follows; each layer runs 4 Mamba (v1) blocks ($d_\mathrm{model}=64$, $d_\mathrm{state}=16$, convolution width 4, expansion 2, RMS normalisation) on the sequence and 4 on the time-reversed sequence, applies a Conv1d (kernel 3) per direction, sums the two directions and adds the layer input; a residual connection also spans the whole stack. A convolutional decoder (Conv1d, kernel 3, batch normalisation) with a skip connection from the encoder output and an output convolution (kernel 3) produces 12 channels. A lead residual adds each normalised input lead to its own output channel, so the network learns a correction for measured channels and a prediction for the others. BiMa-SR has 2,675,148 parameters for $k=1$, 2,675,468 for $k=2$, 2,675,788 for $k=3$ and 2,676,108 for $k=4$.

\subsection{Diffusion formulation}

DiBiMa-SR is a denoising diffusion probabilistic model~\cite{ho2020denoising} with $T=20$ steps, a cosine noise schedule~\cite{nichol2021improved} rescaled to zero terminal signal-to-noise ratio~\cite{lin2024common} and trailing timestep spacing. Its encoder receives the noisy 12-lead target $x_t$ concatenated with the $k$ normalised input leads; a sinusoidal timestep embedding followed by a two-layer MLP is added after the encoder. The network predicts the clean target $x_0$ (``sample'' prediction~\cite{salimans2022progressive}) and is trained with a mean-squared error on $x_0$, with $t$ drawn uniformly from $\{0,\dots,19\}$. Because $\sqrt{\bar\alpha_{19}}=0$, sampling starts from pure Gaussian noise $x_{19}\sim\mathcal{N}(0,I)$ and never from a version of the target. Two samplers were used: one-step, a single $x_0$ prediction at $t=19$, and DDIM-6~\cite{song2021denoising} with $\eta=0$ at $t=19,16,12,9,6,2$. DiBiMa-SR has 2,687,308 parameters ($k=1$). Test-set sampling used a fixed noise seed.

\subsection{Training and model selection}

All reconstruction models were trained in PyTorch~\cite{paszke2019pytorch} for 20 epochs with AdamW~\cite{loshchilov2019decoupled} (learning rate $10^{-3}$ with per-step cosine decay to $10^{-5}$~\cite{loshchilov2017sgdr}; weight decay $10^{-6}$), an effective batch of 32 windows (4 micro-batches of 8), bf16 autocast, gradient-norm clipping at 1 and a mean-squared-error loss on the 12-lead target, with seed 42. After every epoch, the pooled median PCC over the predicted leads was computed on the fixed 1,024 validation windows (DDIM-6 sampling for DiBiMa-SR), and the epoch with the highest value was retained for all test results (Supplementary Fig.~S1). At evaluation, the measured input channels of the output were replaced by the true inputs.

\subsection{Baselines}

The linear filter is a multi-input finite-impulse-response filter: each output lead is a sum of one 101-tap filter (lags $-50$ to $+50$ samples) per input lead, applied to the same normalised inputs as the neural models. The coefficients were fitted jointly for all leads by least squares (float64 normal equations with a ridge term of $10^{-3}$) on 4,000 randomly drawn PTB-XL training windows (seed 0). It has 101 $\times$ (input leads) $\times$ (predicted leads) effective coefficients, that is 1,111 to 3,232 across the six configurations. For configurations containing I and II, ``Einthoven + linear filter'' computes III $=$ II $-$ I, aVR $=-($I $+$ II$)/2$, aVL $=$ I $-$ II$/2$ and aVF $=$ II $-$ I$/2$ exactly from the destandardised inputs~\cite{einthoven1912different,goldberger1942simple} (then z-scored) and takes the chest leads from the linear filter.

\subsection{Reconstruction metrics}

For every window and lead we computed the Pearson correlation coefficient and the mean absolute error between the reconstruction and the stored z-scored target; RMSE is reported in the source data. Only predicted leads were scored, grouped as all predicted leads, predicted limb leads and predicted chest leads. Two PCC statistics were used. The \emph{pooled median PCC} is the median over all window $\times$ lead pairs of a group in a cohort and is the headline accuracy measure; MAE is the mean over the same pairs. The \emph{per-record PCC} is the mean PCC over a record's predicted leads of the group and its windows; it is used for paired comparisons between methods, where each record contributes one difference.

\subsection{Diagnostic-readability classifier}

The classifier is an xresnet1d-style one-dimensional ResNet~\cite{he2016deep,he2019bag,strodthoff2021deep} with kernel size 7, stage widths 64/128/192/256, two residual blocks per stage and a concat-pooling head, with 17 sigmoid outputs (5 superclasses and the 12 panel labels) and 3,364,369 parameters (3.36\,M). It was trained on real 12-lead ECGs of PTB-XL folds 1--8 (record 10565 excluded) with a masked binary cross-entropy, AdamW (learning rate $10^{-3}$, weight decay $10^{-2}$), a one-cycle cosine schedule over 30 epochs (30\% warm-up), batch 128, bf16 autocast, gradient clipping at 1, random 2,250-sample crops and seed 0, with early stopping (patience 5) on the fold-9 record-level superclass macro AUROC. Every input window is z-scored per lead before the network, so real and reconstructed ECGs are processed identically and absolute amplitudes are not visible to the classifier. The trained classifier was applied unchanged to real ECGs and to reconstructed 12-lead ECGs (measured leads re-inserted). A record's probability is the mean over its windows. AUROC was computed per label (Mann--Whitney form~\cite{hanley1982meaning}) on records with a known label; macro AUROC averages the scored labels. PTB-XL superclasses were scored on records with at least one superclass label (2,158 test records). Panel labels with fewer than 10 positive records were skipped (PTB-XL AFLT and WPW; Georgia WPW), and AFLT was never scored in China.

\subsection{Statistics}

Confidence intervals are 95\% percentile intervals of a record-level bootstrap~\cite{efron1979bootstrap,efron1993introduction} with $B=1{,}000$ resamples, in which both windows of a record are resampled together, implemented as weighted statistics with multinomial record counts; the resamples are fixed per cohort (seeds 0, 1 and 2 for PTB-XL test, Georgia and China) and shared by all methods and inputs, so that paired differences are computed on identical resamples. Paired AUROC differences are the mean over resamples of the per-resample difference, with a percentile CI and a two-sided bootstrap $P$ value. For reconstruction accuracy, paired per-record differences are summarised by their mean with a bootstrap CI, the fraction of records in which the first method is better, and a two-sided Wilcoxon signed-rank test~\cite{wilcoxon1945individual}; $P$ values below the floating-point range are reported as $P<10^{-300}$. Changes between PTB-XL and an external cohort use independent bootstraps of the two cohorts.

\subsection{Efficiency measurement}

Inference cost was measured on one workstation with an NVIDIA RTX 4000 Ada Generation GPU (20\,GB) and a 24-core CPU. GPU timings used the \texttt{mamba\_ssm} CUDA kernels with bf16 autocast: latency is the median over 100 timed calls (after 20 warm-up calls) for one 10-s record, that is a batch of its two 5-s windows, and throughput was measured at a batch of 128 windows. CPU timings used a pure-PyTorch implementation of the Mamba selective scan in float32 with 2 and 24 threads (median of 10 timed calls after 3 warm-up calls). The linear filter was timed in NumPy on one CPU thread for all predicted leads and was not timed on the GPU. DDIM-6 requires six network evaluations per window, the other models one.

\section*{Data availability}
All data are public and were obtained from PhysioNet~\cite{goldberger2000physionet}: PTB-XL v1.0.3 (\url{https://physionet.org/content/ptb-xl/1.0.3/})~\cite{wagner2020ptbxl}; the Georgia 12-lead ECG database, part of the PhysioNet/Computing in Cardiology Challenge 2021 training data (\url{https://physionet.org/content/challenge-2021/})~\cite{reyna2021will}; and the Chapman-Shaoxing/Ningbo database, ``A large scale 12-lead electrocardiogram database for arrhythmia study'' v1.0.0 (\url{https://physionet.org/content/ecg-arrhythmia/1.0.0/})~\cite{zheng2022largescale}. All three are distributed under the Creative Commons Attribution 4.0 International licence. 
Source data for every figure, the consolidated results tables and the data quality-control report are provided with the code.

\section*{Code availability}
The code for data preprocessing, model training and evaluation, the statistics and figure scripts, and an interactive demonstration will be made public upon publication in the GitHub repository \url{https://github.com/UgoLomoio/BiMa-ECGSR} and as the Hugging Face Space \url{https://huggingface.co/spaces/Ugo96/BiMa-ECGSR}. 


\section*{Competing interests}
The authors declare no competing interests.

\bibliographystyle{unsrtnat}
\bibliography{biblio}

\end{document}